# Using colorization as a tool for automatic makeup suggestion

Shreyank Narayana Gowda

2017280174

sny17@mails.tsinghua.edu.cn

***Abstract***: *Colorization is the method of converting an image in grayscale to a fully color image. There are multiple methods to do the same. Old school methods used machine learning algorithms and optimization techniques to suggest possible colors to use. With advances in the field of deep learning, colorization results have improved consistently with improvements in deep learning architectures. The latest development in the field of deep learning is the emergence of generative adversarial networks (GANs) which is used to generate information and not just predict or classify. As part of this report, 2 architectures of recent papers are reproduced along with a novel architecture being suggested for general colorization. Following this, we propose the use of colorization by generating makeup suggestions automatically on a face. To do this, a dataset consisting of 1000 images has been created. When an image of a person without makeup is sent to the model, the model first converts the image to grayscale and then passes it through the suggested GAN model. The output is a generated makeup suggestion. To develop this model, we need to tweak the general colorization model to deal only with faces of people.*

## INTRODUCTION

Traditionally, colorization needed user intervention in the form of scribbling of colors over selected regions, or using similar images to extract features for colorization (again a user task as he has to determine a similar image first) or performing preprocessing methods such as image segmentation to obtain segmented parts that can be used for the task of colorization.

Adding color to images can be done for many possible reasons. Some of these include, dormant memories that need to be rekindled, converting age-old black and white movies to color, expressing artistic creativeness etc.

As mentioned above, there are 2 main methods to colorize images:

- User guided colorization
- Automatic colorization without user intervention

User guided colorization often exhibit excellent colorization results. However, it becomes a tedious job. There are many factors that this type of colorization algorithm depends on. Some of these are need of an expert human colorizer to hint to the system, each color region of the image has to be explicitly stated, difficult to determine what shade of color has to be used etc.

Automatic colorization methods usually need little help from the user. They can be trained automatically without need for human intervention. Some of these recent methods have proved to produce some excellent results.

In the next section we discuss in brief about some of the methods in both user guided colorization and automatic colorization. Further, we implement some of these methods and view the results of the same. Lastly, we propose a novel new architecture to automatic colorization and then tweak the same slightly for automatic makeup suggestion.

# BACKGROUND

One of the earliest colorization methods was proposed by Levin et al. [1]. They used user scribbles as input to an image and propagated these scribbles through the image using optimization techniques. A quadratic cost function was proposed and used. This function was used to determine difference between a pixel and its neighbors. Using these differences the algorithm could spread color.

Huang et al. [2] improved this method by preventing color bleeding (method of color extending beyond the necessary boundaries). They used a reliable edge detection method that was quick in speed of execution whilst also being accurate enough to prevent color bleeding.

Chrominance blending was a method suggested by Yatziv et al. [3] to improve the performance of colorization. The chrominance blending was done based on geodesic distances that was weighted. This led to an increase in speed of the algorithm as well.

Luan et al. [4] proposed the use of texture similarity for better performance in colorization. They introduced 2 steps: color labelling and color mapping. Pixels with roughly similar colors are grouped together in coherent regions and color mapping is used to finetune the colorization process.

To aid with propagation of long-range of image recolorization or editing tonal values, An et al. [5] proposed the use of an affinity-based approach. They used global optimization with all-pair constraints.

The above mentioned methods produce some very good results. However, they are heavily dependent on user intervention in their input images. Further, they use trial and error methods to obtain their best results.

To overcome some of these problems, color transfer techniques have been used. These exploit color similarities between a provided reference image and the given input image. One of the earliest color transfer techniques was proposed by Reinhard et al. [6]. They used simple statistical analysis to impose color characteristics of one image over the other.

Pitie et al. [7], provided a method for grading different colors in different images. They found a one-one mapping of colors that transfer from one reference image into the input image directly. For this they developed a mapping algorithm that can transform any N-dimensional probability density function into another.

The above mentioned methods compute statistics related to color information in both the input and a given reference image. Using these statistics, they establish mapping functions that map color distribution of the reference image to the given input image.

Using the color transfer mechanism as a backbone, Welsh et al. [8] proposed a technique to colorize images by matching luminance and texture information between the images. An improvement was observed when a global optimization network which dealt with multi-modal probability prediction was used. The prediction was done for colors at each possible pixel. This method was proposed by Irony et al. [9].

Gupta et al. [10] proposed an approach of matching super pixels. The superpixels were matched between the input image and a given reference image. The matching was done using features extracted from each image and further, space voting was done to obtain colorization results.

Again, the above mentioned results perform well, however, they need a reference image that contains similar information to the given input image. This is a time consuming task and certainly needs user intervention.

Cheng et al. [11], proposed a completely automatic approach for colorization. Various features were extracted from the image. These features were used to colorize patches in the image by utilizing a neural network. In order to improve the results joint bilateral filtering was utilized.

Recent methods have used for superior architectures aided by deep neural networks. Iizuka et al. [12] proposed an architecture that used low-level, mid-level and global features for colorization. They proposed the use of a classification network to aid the colorization process. The classification network shared weights with an autoencoder. Further, the output of the classification network was used as part of a fusion layer with the encoder architecture.

Zhang et al. [13] use the underlying uncertainty of a colorizing problem by first using it as a classification problem and then utilizing class rebalancing at the time of training. This helped to increase the diversity of colors in the result. They implemented a feed forward CNN that was trained on over a million images.

Both semantic representations and low-level feature representation was exploited by Larsson et al. [14]. They trained their model to predict per pixel color histograms. They do this because most images have scene elements that can be represented by multi-modal representations.

Guadaramma et al. [15] proposed an approach based on the idea that it is easy to colorize an image if we train a model with low resolution color images. To do this they train a conditional PixelCNN [16] to generate a low resolution color image for a given grayscale image. They then train another CNN to convert this low resolution image into a high resolution image.

Makeup suggestion is a relatively new field in computer vision. It is the task of taking an input image of a person and suggesting the possible makeup for that person based on certain features in the person.

There are very few datasets available for this task. Chen et al. [17] have provided one such dataset, but for a different context. This was provided for detecting makeup on a person, but it is a dataset from which certain images can be extracted for the purpose of makeup generation.

Dhall et al. [18] suggest an adaptive makeup algorithm that is automatic in nature. It applies makeup based on the ethnicity of skin color and gender of the person in the image. However, this algorithm gives the same results for people having same skin color and does not look specifically for facial features. However, one big advantage is that the process was completely automatic and did not need user intervention.

A hardware based system was developed by Iwabuchi et al. [19] to make the makeup process easier to do. Facilities like automatic zoom to a particular part of the face, displaying face from different angles etc was provided with the aid of cameras in certain positions of an electronic dressing table. However, this is a very expensive process to set up.

Some of the recently developed methods for automatic makeup suggestion are as follows. Li et al. [20] proposed the separation of the input image into various intrinsic layers that can be modified based on certain reflectance models. They believed this would aid in generating realistic results without the need for detailed geometric and reflectance measurements of the user.

Xu et al. [21] provided an automatic method to suggest virtual makeup to a facial image based on example face images with makeup. They detected face landmarks and adjusted the landmarks based on skin color Gaussian Mixture model based segmentation. The skin layer was further separated into three layers and makeup is transferred to each layer.

Alashkar et al. [22] developed a system that classifies makeup related facial traits that professional artists consider key to a makeup style. They then use a rule-based makeup recommendation system by developing a knowledge base system that models relationship between facial attributes and makeup attributes. They take into account the level of makeup needed and the desired trend.

Another fully automatic approach was proposed by Alashkar et al. [23]. Here, they used three stages for suggestion of makeup. Firstly, they classified facial traits into structured coding. Then, they passed these traits as input into a rule guided deep network model that utilizes pairwise before-after makeup images and knowledge of a professional makeup artist. Finally, they provide visual results of their system.

We propose a colorization approach to makeup suggestion. The colorization process automatically suggests color to facial traits that have makeup on based on the training done by the model. The model's architecture, the proposed dataset and results are shown in the following sections. First, however, we look at the results of some of the executed models from previous papers.

## EVALUATED ALGORITHMS FOR COLORIZATION

The first model implemented was the model developed in [12,31]. The architecture of the model taken from the paper is shown in Figure 1.

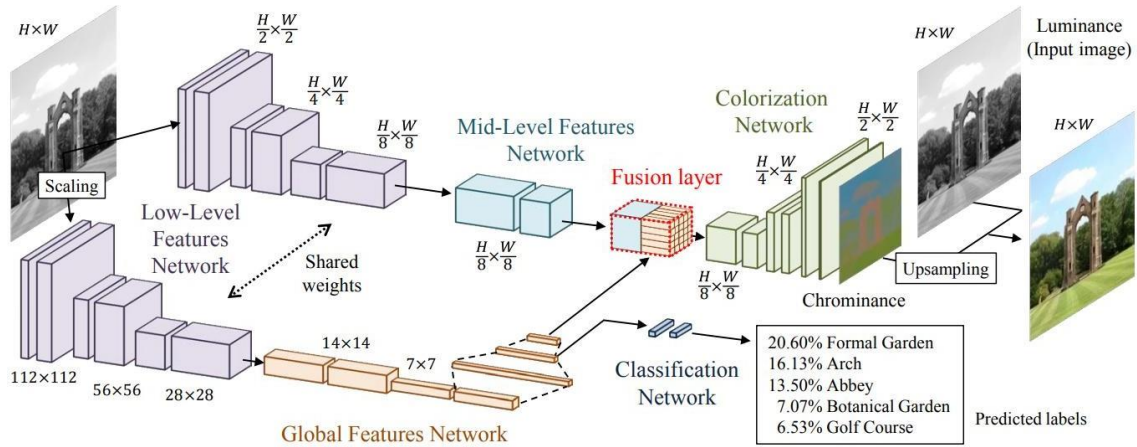

**Figure 1. Architecture of model in [12]**

The model was developed in Tensorflow from scratch. The specifications of the model are exactly the same as those provided in the paper. Figure 2 shows some of the outputs obtained from the model.

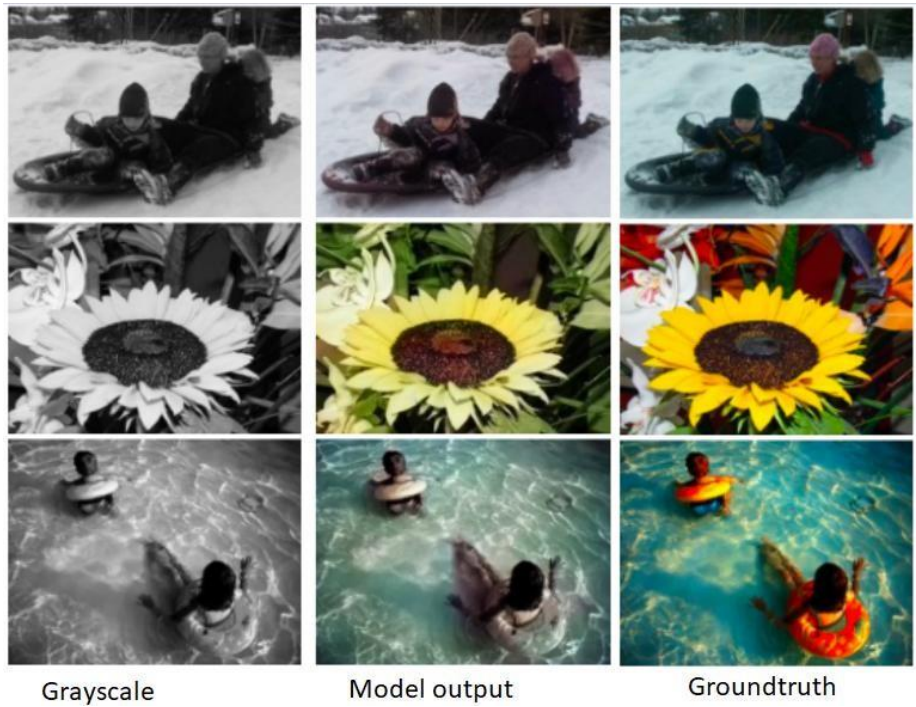

**Figure 2. Output from model in [12]**

The next model implemented was not a paper. However, it is a model that formed the basis for many of the recent papers. This model was developed with the belief the more information can be obtained from a pretrained CNN than just classification results. This model is shown in Figure 3 and was used with residual connections.

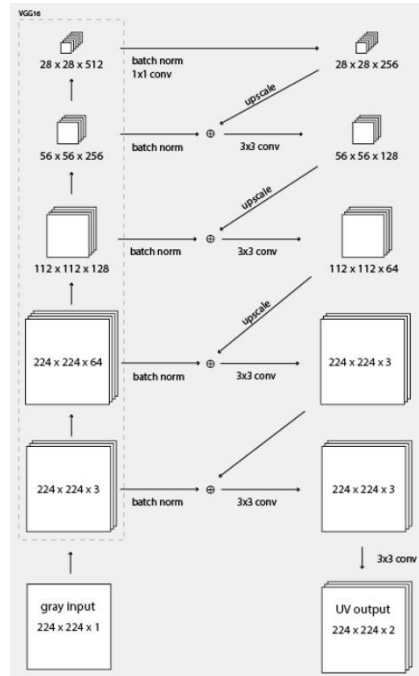

**Figure 3. Architecture of residual model**

The results obtained from this model are shown in Figure 4.

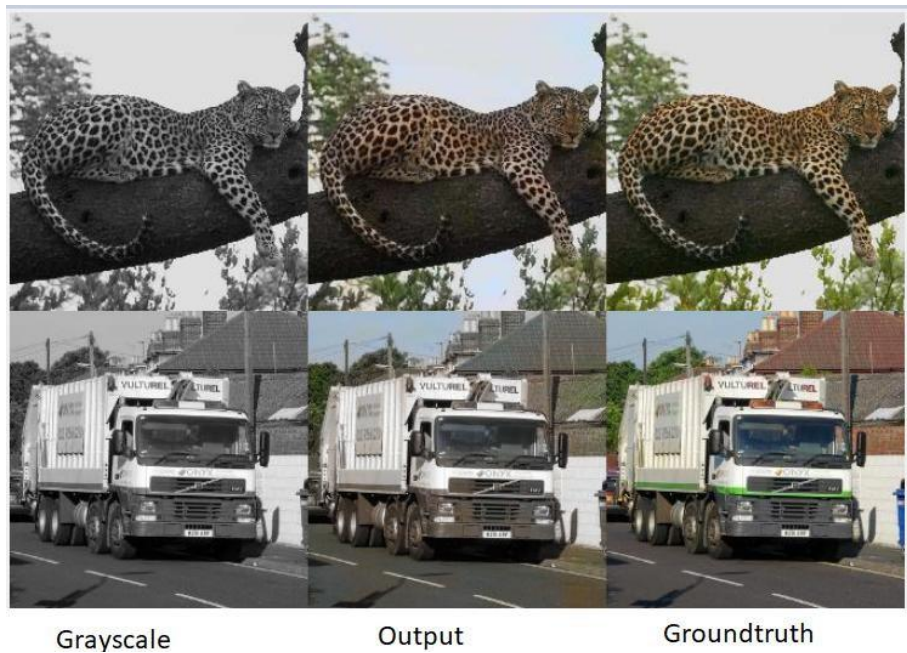

**Figure 4. Output of residual model**

We look at the model proposed in [12] and believe that the classification network can be replaced with pretrained models and a simple autoencoder used can be replaced with GANs for more realistic looking results. The proposed architecture includes the same. The proposed method is discussed in the next section.

# PROPOSED METHOD

## MODEL ARCHITECTURE

As mentioned in the previous section a few tweaks have been made into the architecture of the model proposed in [12]. First, we believe that the classification network used is very simple and that using a better network pretrained on a big dataset like imagenet [24] would give better results.

Pretrained models such as VGG [25], Inception-Resnet [26], Resnet [27], Densenet [28]etc are available. These models have the ability to classify the contents of the image more accurately than the simple network used in the model.

Further, utilizing GANs [29] over autoencoders for training should yield better results because of the ability of the GAN to generate better images. In particular we consider the use of DC-GANs proposed by Radford et al. [30] as the model for colorization. This is because the training of GANs can be unstable and difficult to deal with.

This proposed model is a general colorization model. The classification network is removed when we want to develop an automatic makeup suggestion system. This is because the makeup system deals with only face images and we do not need a model pretrained on imagenet for the same.

We use the LAB color space instead of the RGB color space because of the minimization of per-pixel correlation in the LAB color space. The L channel called the lightness channel is essentially a grayscale version of the color image and the AB channels are chrominance channels that add color to the image.

Formally, given $I_L$ which is a grayscale image we wish to generate the chrominance value $I_{AB}$ and combine it with $I_L$ to obtain our final colorized image.

We explore the possibility of using individual channel prediction and hence use two DC-GANs, one for A channel and one for B. The training of the models follows the same basic process, only the input and output are different. The architecture model for general colorization is depicted in Figure 5.

We use a DenseNet with growth factor 48 and depth of 161 as our pretrained model as we believe it exhibits state of the art results on ImageNet dataset.

The training procedure is explained in detail in the following section. Details about the model and loss function are also stated.

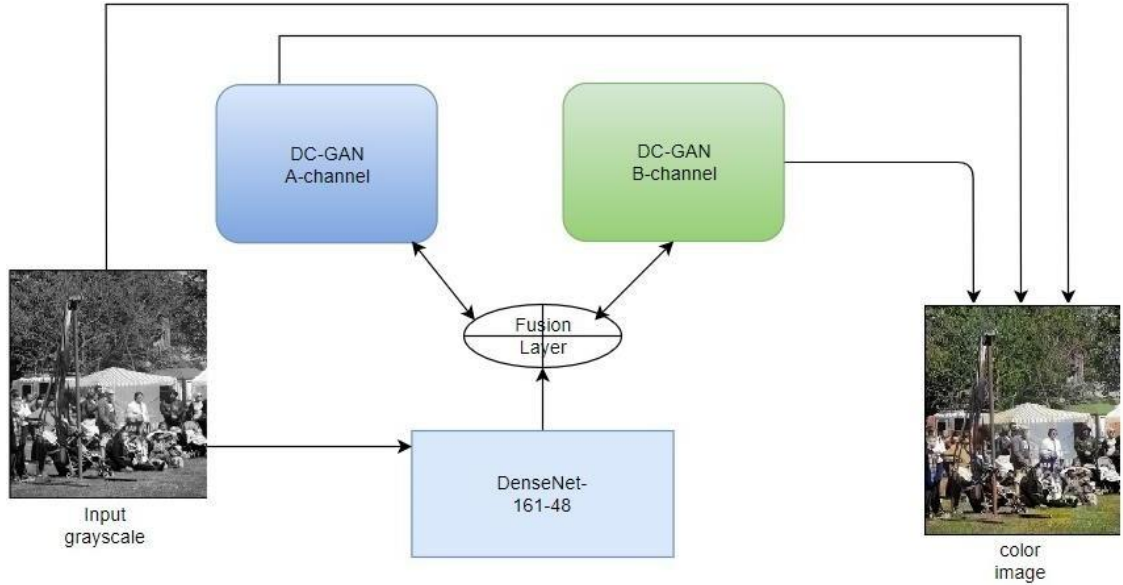

**Figure 8. Model Architecture**

## MODEL TRAINING

We train generator models in the DC-GAN for both A and B channels. Essentially the generator model tries to generate a mapping function between the image in grayscale say $X_L$ with either the A channel or B channel depending on the generator. Let 'p' refer to the vector of components of $p^{th}$ pixel in the image and 'c' refers to the channels in the image: in this case 'A' and 'B'. 'n' refers to the total number of pixels in the image. G is a generator model which maps the given channel with the L channel. We want to minimize the least squares objectives shown in (1).

$$L(X, \emptyset) = \frac{1}{n} \sum_{p=1}^{n} ||G(X_L; \emptyset)^{(p)} - X_C^{(p)}|| \quad (1)$$

We further use minibatch stochastic gradient with momentum as our optimizing function. Momentum is represented by 'μ' and learning rate by 'λ'. We define our minibatch loss as the average of individual example losses. This is shown in (2).

$$L_B = \frac{1}{X_B} \sum_{X \in X_B} L(X, \emptyset) \quad (2)$$

We use ReLu functions for activating the nodes in the network. Additionally, the generator G adds noise 'z' as input. We also build a discriminator model D which takes as input the grayscale image along with the colorized results from the respective generators. The output from the discriminator is a prediction of the probability that the generated image is as close to the original ground truth color image as possible. The discriminator is trained to detect results that are not good enough by the generator.

For each DC-GAN we are trying to minimize the losses from both discriminator and from the generator. We use cross entropy loss for the same. This can be seen in (3).

$$-logD(X_C|X_L) - log(1 - D(G(z|X_L)|X_L))$$

## MODEL SPECIFICATIONS

Table 1 shows how small blocks are defined in the model.

| Block Name | Contents |
| --- | --- |
| Block A | 2 Conv layers and 1 Max pool layer |
| Block B | 2 Conv layers, 1 upsample layer and 1 Merge layer that merges output of conv layer with upsample layer |
| Block C | 3 Conv layers |

Each generator in the proposed model has 4 Block A blocks, 4 Block B blocks and 1 Block C block. Each discriminator model contains 4 small blocks of alternating convolutional layers and max pool layers. This is followed by a single convolutional layer, one flatten layer and one dense layer.

## DATASET

For the general colorization purpose we trained and tested on the imagenet dataset. We used 600000 randomly selected images for training and a further 10000 for testing purposes. The results of the same are shared in the next section.

For the makeup suggestion model we develop a dataset of 1000 images. Each image is a color image that has been cropped to contain only the main facial features for which we can apply makeup. Each image is 256x256 in size. A sample set of images in the dataset is displayed in figure 9. The dataset was developed using google, flickr and other such publicly available images.

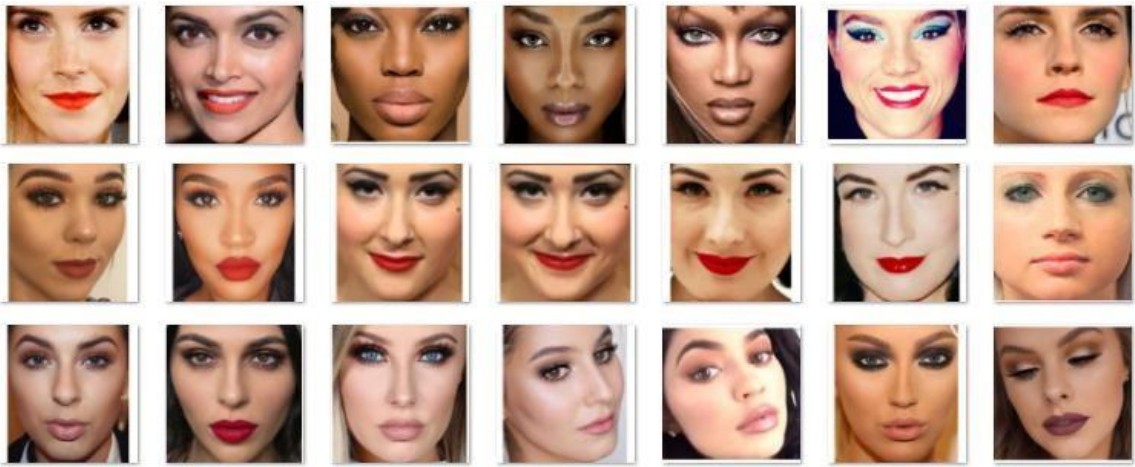

Figure 9. Sample images set

Each image was manually cropped so as to retain the most important parts of the face. Ethnicity was also given importance whilst choosing the images.

## EXPERIMENTAL RESULTS

Figure 10 contains results for the general colorization model.

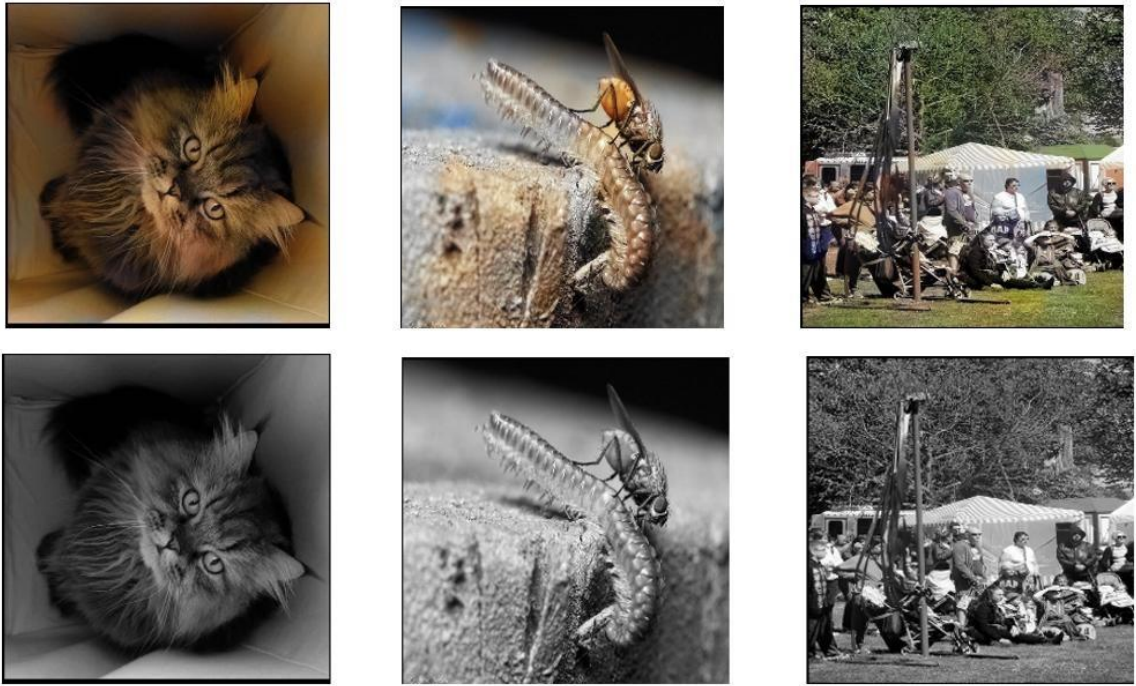

**Figure 10. Sample results from model**

Figure 11 contains some of the results for makeup generation.

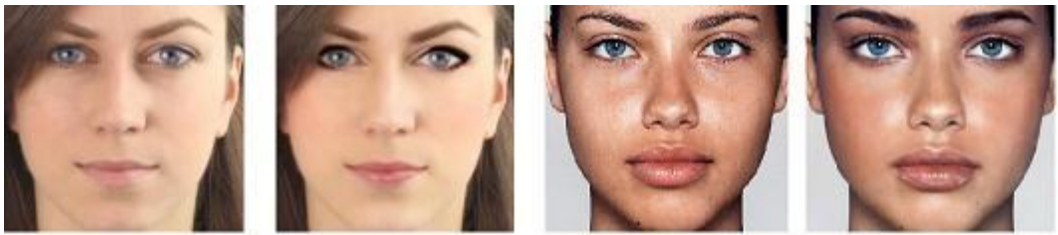

**Figure 11. Sample makeup generated results**